%% file: iclr2025_conference.tex
\documentclass{article} 
\usepackage{iclr2025_conference,times}

\input{math_commands.tex}

\usepackage{hyperref}
\usepackage{url}
\usepackage{amsmath}
\usepackage{placeins}
\usepackage{amsfonts}
\usepackage{amsthm}
\usepackage{graphicx}
\usepackage{textcomp}
\usepackage{xcolor}
\usepackage{listings}
\usepackage{float} 
\usepackage{pythonhighlight}
\usepackage[T1]{fontenc}
\usepackage{verbatim}
\usepackage{url}
\usepackage{multirow}
\usepackage{booktabs} 
\usepackage{wrapfig}
\usepackage{tcolorbox}
\usepackage{algorithm}
\usepackage{algpseudocode}
\usepackage{amsmath}
 
\usepackage{float}
\usepackage{enumitem}
\usepackage{caption}
\usepackage{caption}
\usepackage{subcaption}
\usepackage{afterpage}
\usepackage{amssymb}  
 
\usepackage{xcolor}   
\usepackage{soul} 
\usepackage{xcolor} 
\sethlcolor{yellow} 
\usepackage{booktabs}       
\usepackage{amsfonts}       
\usepackage{nicefrac}       
\usepackage{microtype}      
\usepackage{graphicx}       

\usepackage{enumitem}
\usepackage{colortbl}       
\usepackage[utf8]{inputenc}
\usepackage[T1]{fontenc}
\usepackage{graphicx}%
\usepackage{mathtools}
\usepackage{multirow}%
\usepackage{amsfonts}%
\usepackage{amsthm}%
\usepackage{mathrsfs}%
\usepackage[title]{appendix}%
\usepackage{xcolor}%
\usepackage{textcomp}%
\usepackage{manyfoot}%
\usepackage{booktabs}%
\usepackage{algorithm}%
\usepackage{algorithmicx}%
\usepackage{algpseudocode}%
\usepackage{listings}%
\theoremstyle{plain}
\newtheorem{theorem}{Theorem}[section]

\newtheorem{lemma}[theorem]{Lemma}
\newtheorem{corollary}[theorem]{Corollary}
\theoremstyle{definition}

\theoremstyle{remark}

\usepackage{lineno}
\usepackage{xcolor}
\usepackage{pifont}
\definecolor{waymollgreen}{HTML}{CCFAEB} 
\definecolor{waymoblue}{HTML}{0077FF} 

\title{Efficient Temporal Consistency in Diffusion-Based Video Editing with Adaptor Modules: A Theoretical Framework}

\author{
  \begin{tabular}{c}
    Xinyuan Song$^{1}$, Yangfan He$^{2}$, Sida Li$^{3}$, Jianhui Wang$^{4}$,\\ Hongyang He$^{5}$,
    Xinhang Yuan$^{6}$, Ruoyu Wang$^{7}$, Jiaqi Chen$^{8}$,\\
    Keqin Li$^{8}$, Kuan Lu$^{9}$, Menghao Huo$^{10}$, Binxu Li$^{11}$, Pei Liu$^{12}$ \\[0.5em]
    $^{1}$ Emory University, 
    $^{2}$ University of Minnesota—Twin Cities,\\ 
    $^{3}$ Peking University,
    $^{4}$ University of Electronic Science and Technology of China,\\
    $^{5}$ University of Warwick, 
    $^{6}$ Washington University, Saint Louis,
    $^{7}$ Tsinghua University, \\
    $^{8}$ Independent Researcher, 
    $^{9}$ Cornell University,
    $^{10}$ Santa Clara University,\\ 
    $^{11}$ Stanford University, 
    $^{12}$ Hong Kong University of Science and Technology
  \end{tabular}
}

\iclrfinalcopy 
\begin{document}

\maketitle

\begin{abstract}
Adapter-based methods are commonly used to enhance model performance with minimal additional complexity, especially in video editing tasks that require frame-to-frame consistency. By inserting small, learnable modules into pretrained diffusion models, these adapters can maintain temporal coherence without extensive retraining. Approaches that incorporate prompt learning with both shared and frame-specific tokens are particularly effective in preserving continuity across frames at low training cost. In this work, we want to provide a general theoretical framework for adapters that maintain frame consistency in DDIM-based models under a temporal consistency loss. First, we prove that the temporal consistency objective is differentiable under bounded feature norms, and we establish a Lipschitz bound on its gradient. Second, we show that gradient descent on this objective decreases the loss monotonically and converges to a local minimum if the learning rate is within an appropriate range. Finally, we analyze the stability of modules in the DDIM inversion procedure, showing that the associated error remains controlled. These theoretical findings will reinforce the reliability of diffusion-based video editing methods that rely on adapter strategies and provide theoretical insights in video generation tasks.
\end{abstract}

\section{Introduction}
From natural language processing , to time series analysis  and computer vision , deep learning has made significant strides across multiple disciplines~\cite{qiu2025easytime,  qiu2025duet, qiu2024tfb, tao2024robustness,du2025zero,shen2024altgen,wang2025research, wang2025research, wang2025design, zhao2025optimizedpathplanninglogistics, xu2024autonomous, xu2024comet, weng2022large,zhong2025enhancing,li2025revolutionizing, li2024towards, li2023bilateral, li2024distinct}. And text-to-image (T2I) diffusion models~\cite{liu2024sorareviewbackgroundtechnology, wang2023modelscopetexttovideotechnicalreport, khachatryan2023text2videozero, bartal2024lumierespacetimediffusionmodel, Tune-A-Video} have brought significant progress to the field of generative image modeling. Building on their success, text-to-video (T2V) frameworks aim to preserve temporal consistency across frames. However, many T2V solutions involve considerable training overhead or extensive parameter sizes. To reduce these challenges, researchers have proposed fine-tuned T2I adapters for video editing~\cite{mou2023t2iadapterlearningadaptersdig, balaji2023ediffitexttoimagediffusionmodels, chen2023visiontransformeradapterdense, feng2023trainingfreestructureddiffusionguidance}. A frequently used strategy is to enforce a cosine similarity constraint between feature maps of adjacent frames during early denoising steps~\cite{geyer2023tokenflow}, in addition to a binary cross-entropy objective on noise prediction. Low-Rank (LoRA) Adaptation on cross-attention layers helps limit parameter growth, and shared prompt tokens further promote frame-to-frame coherence.

In this paper, we offer a detailed theoretical analysis of these techniques, with particular emphasis on the temporal consistency loss and the DDIM-based framework. In Theorem~\ref{thm:temporal_consistency}, we establish that the popular temporal consistency loss is differentiable under bounded feature norms, and its gradient is Lipschitz continuous. This implies that gradient-based methods can optimize the loss without risk of divergence. We then demonstrate in Theorem~\ref{thm:temporal_convergence} that standard gradient descent will monotonically reduce the loss and converge to a local minimum, provided the step size is suitably chosen. Further, Lemma~\ref{lemma:temporal_convexity} shows that the temporal consistency objective can be viewed as a convex function of the inter-frame similarity terms, revealing a favorable optimization landscape.

We also address the DDIM inversion step, where one typically seeks to refine the latent representations of consecutive frames. Theorem~\ref{lemma:bilateral_stability} proves that incorporating bilateral filtering leads to a bounded error that does not escalate through successive iterations of reverse diffusion. The analysis involves characterizing how the filtering operator contracts errors and how new noise injection influences the total error. This property ensures the stability of the video editing process and prevents divergence even when multiple frames are processed in sequence.

Lastly, we show how shared and unshared tokens in the prompt-learning stage can theoretically approximate any desired feature representation, if their numbers are large enough relative to the feature dimension. Corollary~\ref{cor:token_sufficiency} states that once the token embedding space can span the entire feature dimension, the alignment error of the cross-attention output can be driven to arbitrarily small values. This result underlines the flexibility of prompt-based adapters in capturing nuanced frame dependencies.

Our theoretical insights show why these adapter-based strategies are stable, convergent, and capable of improving temporal consistency without increasing model size excessively. By ensuring bounded errors, convex objectives, and sufficient expressive power in the token embeddings, the proposed approach can reliably generate coherent video sequences from pretrained T2I diffusion models.

We conducted extensive empirical studies to validate our theoretical findings. Specifically, our experiments systematically analyzed the impact of the UNet adapter, demonstrating its effectiveness in maintaining consistent structural and semantic coherence across consecutive frames. Furthermore, quantitative evaluations using cosine similarity metrics corroborate that our theoretical predictions hold true empirically: models equipped with the adapter consistently exhibit higher temporal coherence, achieving cosine similarity values approaching unity. These findings strongly support our theoretical analysis, confirming that the adapter-based methods indeed promote temporal stability, convergence, and consistency in practice.

The main contributions of this paper are as follows:
\begin{itemize}[left = 0em]
    \item We provide a detailed theoretical analysis of temporal consistency loss, demonstrating its differentiability under bounded norms, Lipschitz continuous gradient, and favorable optimization landscape (convexity in inter-frame similarity terms).
    \item We theoretically establish convergence properties of gradient-based methods optimizing the temporal consistency loss, ensuring monotonic reduction and convergence to a local minimum with appropriate step sizes.
    \item We rigorously analyze the stability of DDIM inversion integrated with bilateral filtering, proving bounded error propagation across successive diffusion iterations, thus ensuring stability in sequential video editing.
    \item We demonstrate theoretically that shared and unshared prompt tokens have sufficient expressive power to approximate arbitrary feature representations, supporting flexible and robust frame-to-frame dependency modeling.
    \item Extensive empirical studies validate our theoretical findings, confirming the effectiveness of the UNet adapter in maintaining structural and semantic coherence across frames, as evidenced by significantly improved cosine similarity and temporal stability metrics.
\end{itemize}

Through these contributions, we aim to advance the field of T2V generation and editing, providing a robust and efficient framework that can leverage the strengths of T2I models while mitigating their limitations.

\section{Related Work}
\label{sec:related_work}
\noindent\textbf{Text-to-Video Editing.}
Text-to-Image (T2I) generation has seen rapid progress, particularly through advances in Generative Adversarial Networks (GANs)\cite{shen2023pbsl, NEURIPS2021_076ccd93, shen2023git, Karras_2020_CVPR, sauer2022styleganxlscalingstyleganlarge} and diffusion-based frameworks\cite{shen2024imagdressing, pan2018createtellgeneratingvideos, soviany2019imagedifficultycurriculumgenerative, shen2023advancing, shen2024boosting, gao2024exploring}. T2V techniques have been categorized into several strategies, including diffusion model inversion and sampling~\cite{rombach2022high, lu2022dpm, shen2024imagpose, salimans2022progressive}, lightweight fine-tuning using adapters~\cite{kim2024arbitrary}, and post-hoc temporal consistency enhancement via attention-based mechanisms. Representative T2V models such as Gen-L-Video~\cite{wang2023gen}, FLATTEN~\cite{cong2023flatten}, and StableVideo~\cite{chai2023stablevideo} aim to improve long-range temporal alignment, while methods like ControlVideo~\cite{zhao2023controlvideo} and MagicProp~\cite{yan2023magicprop} target frame-level fidelity. Building on these advances, works extend diffusion models to text-to-video editing by incorporating parameter-efficient adapters~\cite{xin2024vmt,xin2024mmap,xin2023self}.\\

\noindent\textbf{DDIM Inversion for Video Generation.}
Denoising Diffusion Implicit Models (DDIM) enable latent trajectory manipulation through their invertible structure~\cite{song2022denoisingdiffusionimplicitmodels}. Recent developments like EasyInv~\cite{zhang2024easyinvfastbetterddim} and ReNoise~\cite{garibi2024renoiserealimageinversion} iterate between forward and backward noise steps to improve reconstruction. Time-varying inversion schedules, such as Eta Inversion~\cite{kang2024etainversiondesigningoptimal}, provide enhanced diversity by modulating noise injection spatially and temporally. Additional strategies, including MasaCtrl~\cite{cao2023masactrltuningfreemutualselfattention} and Portrait Diffusion~\cite{liu2023portraitdiffusiontrainingfreeface}, further refine inversion through attention-based noise representations and key-value feature alignment. In vision tasks, modular adapters have facilitated parameter-efficient fine-tuning in architectures such as T2I-Adapter~\cite{mou2023t2iadapterlearningadaptersdig} introduce task-specific guidance into diffusion models. Moreover, Uni-ControlNet~\cite{zhao2023unicontrolnetallinonecontroltexttoimage} proposes a unified approach to integrate control signals across multiple scales. Most importantly, ~\cite{he2025enhancing} propose a General and Efficient Adapter integrating temporal, spatial, and semantic consistency modules with DDIM inversion to significantly improve perceptual quality and temporal coherence for text-to-video editing.

Although numerous recent works explore video-level consistency within diffusion-based frameworks, a mature theoretical foundation for this area is still lacking. In particular, the theoretical understanding of consistency losses and DDIM-based adapters remains underdeveloped. This work aims to address this gap by establishing formal analyses to support principled video editing with diffusion models.

\begin{figure*}[htbp]
    \centering
    \includegraphics[width=\textwidth]{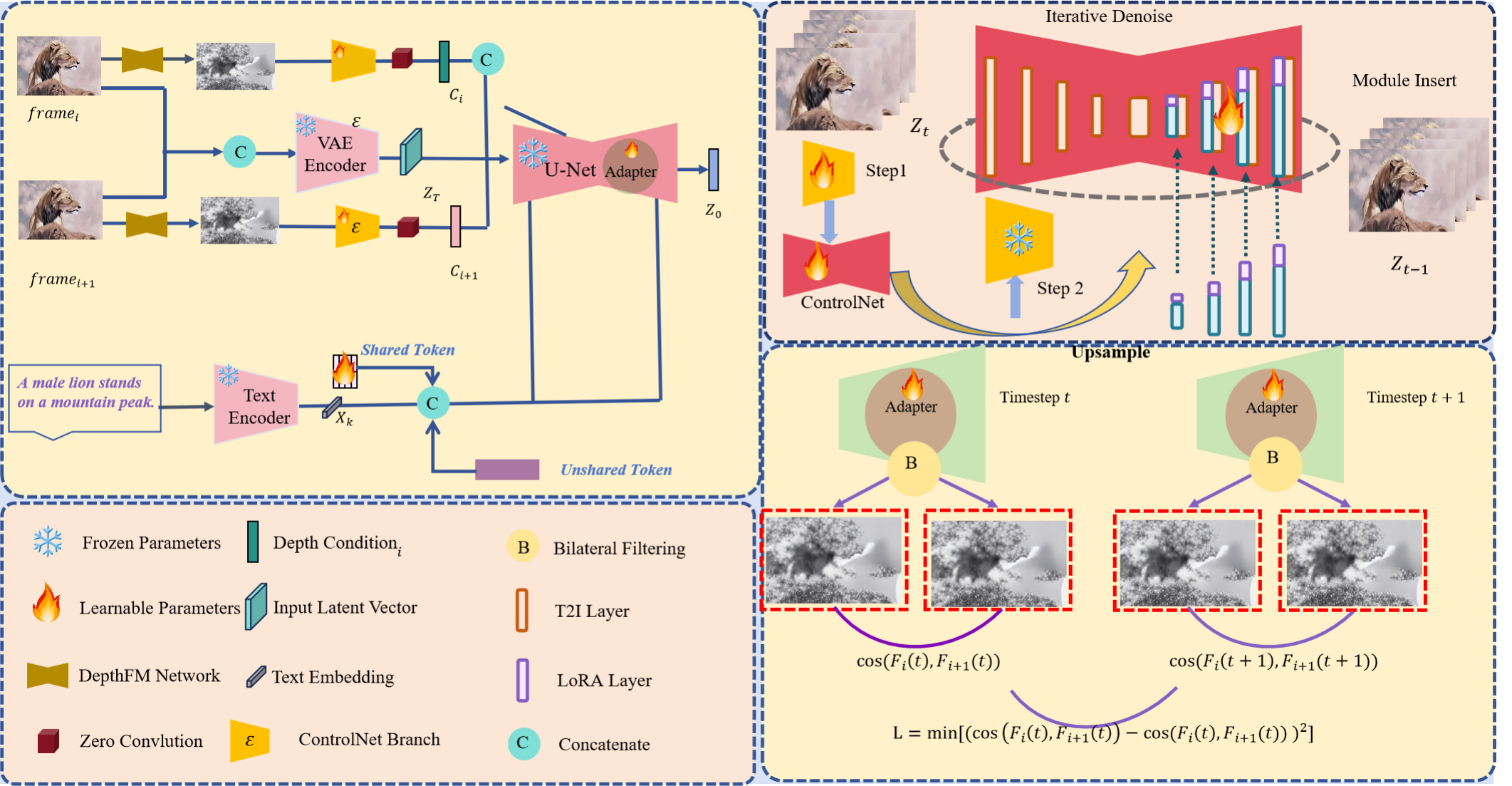} 
    \caption{An overview of the typical video generation process using LoRA-enhanced feature extraction. Depth and text embeddings are combined with latent vectors and processed through iterative denoising, cross attention, and cosine similarity constraints between adjacent frames.}
    \label{fig:lora_video_generation}
    \vspace{-1em}
\end{figure*}

\section{Preliminaries}
\subsection{Diffusion model for video generation} \label{backbone}
The diffusion module~\cite{khachatryan2023text2videozero,qi2024deadiffefficientstylizationdiffusion, tokenflow2023,zhang2023addingconditionalcontroltexttoimage,fraikin2023trep} first encodes consecutive frames ($\mathrm{frame}_i$, $\mathrm{frame}_{i+1}$) into latent representations ($z_t$, $z_{t+1}$) using a VAE. Gaussian noise is added at each diffusion timestep to produce $\mathrm{noise}_1$ and $\mathrm{noise}_2$, applied separately to each frame to capture frame-specific variations. The latent representations and their noisy versions ($[z_t, z_{t+1}, z_t, z_{t+1}]$) support cross-frame temporal modeling. Time embedding vectors encode the timestep $t$, and at each step, the latent are concatenated with control signals ($c_t, c_{t+1}, c_t, c_{t+1}$). Temporal feature maps ($F_t$, $F_{t+1}$) are then injected into the UNet. This combined latent representation enhances spatial-temporal dependencies between frames. Figure~\ref{fig:lora_video_generation} provides an overview of the standard video generation process, illustrating the practical application of our theoretical analysis.

\subsection{Frame Similarity-based Temporal-Spatial Consistency Module}

For the decoder layers:

\begin{equation}
x_{t+1} = x_t + \epsilon_t - \theta(x_t, t),
\end{equation}
where $x_t$ is the image at timestep $t$, $\epsilon_t$ is the predicted noise, and $\theta$ is the UNet model. popular methods incorporate trainable adapters into the UNet, extracting intermediate feature maps $\mathbf{F}_{l,b}^{t}$ from each block $(l, b)$ at timestep $t$:
\begin{equation}
\mathbf{F}_{l,b}^{t} = \mathbf{W}_0 \mathbf{x} + \mathbf{B}_{l,b} \mathbf{A}_{l,b} \mathbf{x},
\end{equation}
where $\mathbf{x}$ is the input feature, and $\mathbf{B}_{l,b}$ and $\mathbf{A}_{l,b}$ are learnable low-rank parameters.

Popular methods use a similarity function to measure alignment between adjacent feature maps:
\begin{equation}
\text{Sim}(\mathbf{F}_t, \mathbf{F}_{t+1}) = \frac{\mathbf{F}_t \cdot \mathbf{F}_{t+1}}{\|\mathbf{F}_t\| \|\mathbf{F}_{t+1}\|}.
\end{equation}
So the temporal consistency loss is defined as:
\begin{equation}
\mathcal{L}_{\text{temporal}} = \frac{1}{T-1} \sum_{t=1}^{T-1} \Big( \text{Sim}(\mathbf{F}_t, \mathbf{F}_{t+1}) - \text{Sim}(\mathbf{F}_{t-1}, \mathbf{F}_t) \Big)^2,
\end{equation}
where $T$ is the total number of timesteps. 

A standard diffusion loss is also required:
\begin{equation}
\mathcal{L}_{\text{diffusion}} = \mathbb{E}_{x_0, \epsilon, t} \Big[ \| \epsilon - \epsilon_\theta(x_t, t) \|^2 \Big],
\end{equation}
where $\epsilon$ is the noise added to $x_0$ at timestep $t$, and $\epsilon_\theta$ is the model’s predicted noise. The overall objective function combines the temporal consistency and diffusion losses:
\begin{equation}
\mathcal{L}_{\text{total}} = \lambda_{\text{temporal}} \mathcal{L}_{\text{temporal}} + \lambda_{\text{diffusion}} \mathcal{L}_{\text{diffusion}},
\end{equation}
where $\lambda_{\text{temporal}}$ and $\lambda_{\text{diffusion}}$ are set to 1 and 0.01, respectively. 

In DDIM inversion for video generation, the reverse diffusion process denoise an input $ x_t $ according to:
\begin{equation}
p_\theta(x_{t-1} | x_t) = \mathcal{N}\bigl(x_{t-1}; \mu_\theta(x_t, t), \Sigma_\theta(t)\bigr),
\end{equation}
where $ \mu_\theta(x_t, t) $ is the predicted mean, and $ \Sigma_\theta(t) $ is the variance schedule. The denoising of a noisy input $ x_t $ is governed by:  
\begin{equation}
x_{t-1} = \frac{1}{\sqrt{\alpha_t}} \left( x_t - \frac{1 - \alpha_t}{\sqrt{1 - \bar{\alpha}_t}} \epsilon_\theta(x_t, t) \right) + \sqrt{1 - \alpha_{t-1}} z,
\end{equation}
which is further refined by applying bilateral filtering to the latent representation~\cite{luo2024enhancing}.
While $ \epsilon_\theta(x_t, t) $ predicts the noise, and $\alpha_t$, $\bar{\alpha}_t$ are scaling factors with $ z $ as sampled noise. We use a typical framework with bilateral filtering step applied to the noisy latent $ x_t $:
\begin{equation}
O_x = \frac{\sum_{y \in \mathcal{N}(x)} G_{\text{spatial}}(x, y) G_{\text{intensity}}(I_x, I_y) I_y}{\sum_{y \in \mathcal{N}(x)} G_{\text{spatial}}(x, y) G_{\text{intensity}}(I_x, I_y)},
\label{eq : video_inv}
\end{equation}
where $ \mathcal{N}(x) $ denotes the neighborhood of pixel $ x $, with $ y $ as neighboring pixels, defined by their respective intensities $ I_x $ and $ I_y $. The spatial and intensity weights are calculated by:
\begin{equation}\label{eq:10}
G_{\text{spatial}}(x, y) = \exp\left(\frac{-(x - y)^2}{2\sigma_{\text{spatial}}^2}\right),
\end{equation}
\begin{equation}\label{eq:11}
G_{\text{intensity}}(I_x, I_y) = \exp\left(\frac{-(I_x - I_y)^2}{2\sigma_{\text{intensity}}^2}\right),
\end{equation}
where $ \sigma_{\text{spatial}} $ determines sensitivity to spatial distances, and $ \sigma_{\text{intensity}} $ controls the filter's response to intensity differences. This framework, in reference to~\cite{he2025enhancing}, is both intuitive and general, as typical existing methods incorporate a similar processing step to enhance the quality and coherence of latent representations.

At each timestep, producing refined latents $ x_t' $. The updated inversion step is:
\begin{equation}\label{eq:DDIM}
x_{t-1} = \frac{1}{\sqrt{\alpha_t}} \left( x_t' - \frac{1 - \alpha_t}{\sqrt{1 - \bar{\alpha}_t}} \epsilon_\theta(x_t', t) \right) + \sqrt{1 - \alpha_{t-1}} z,
\end{equation}
where $ x_t' $ is the filtered latent obtained from $ x_t $, ensuring smoother and more consistent intensity distributions.


\subsection{Shared and unshared Token-Based Consistency Analysis}\label{sec:3}

\begin{figure*}[htbp]
        \centering
        \includegraphics[width=\textwidth]{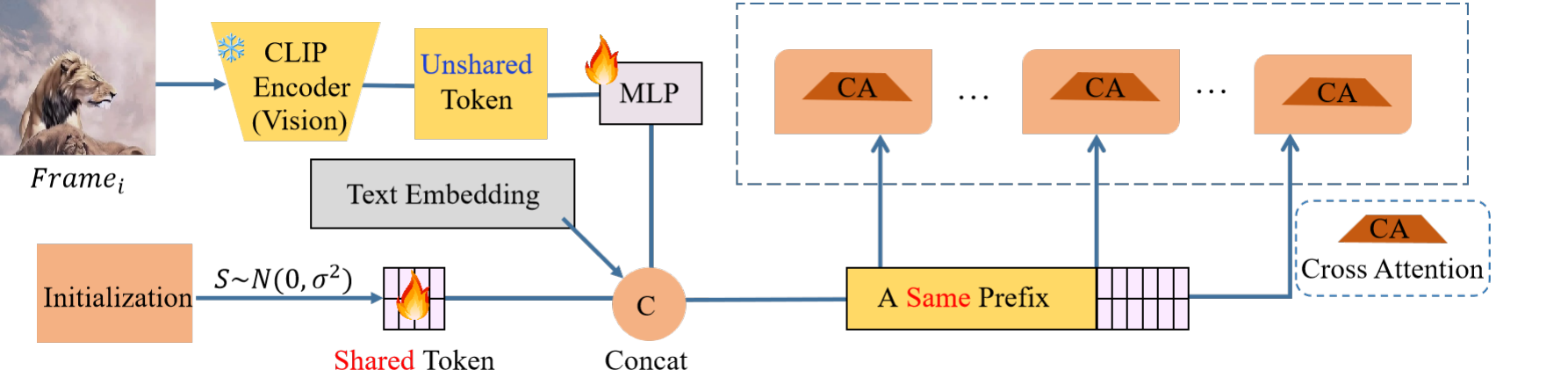} 
        \caption{Typical adapters mechanism regarding shared and unshared token mechanism for video generation. Shared tokens ensure global consistency across frames, while unshared tokens handle frame-specific details. A same prefix is applied across all time steps, and only shared tokens are updated during the final phase.}
        \label{fig:model_performance}
        \vspace{-1em}
\end{figure*}

Figure~\ref{fig:model_performance} illustrates the typical operation of a consistency module employing shared and unshared tokens. The text embedding for temporal-aware fine-tuning is constructed as:
\begin{equation}
Z_{\text{final}} = \left[ T_{\text{share}}; Z_{\text{frame}}; \mathcal{C}(Z) \right],
\end{equation}
where $ T_{\text{share}} $ represents the shared token embedding, $ Z_{\text{frame}} $ is the frame-specific unshared token, and $ \mathcal{C}(Z) $ concatenates conditional and unconditional embeddings along the first sequence dimension. 

During the denoising process, cross-attention provides text guidance by mapping the latent features $ X_t \in \mathbb{R}^{M \times d} $ to updated features $\tilde{X}_t$ using the final text embedding $ Z_{\text{final}} \in \mathbb{R}^{L \times d} $ as keys and values:
\begin{equation}
Q = W_Q^\top X_t,\quad K = W_K^\top Z_{\text{final}},\quad V = W_V^\top Z_{\text{final}},
\end{equation}
\begin{equation}\label{eq:18}
\tilde{X}_t = \text{softmax}\left(\frac{QK^\top}{\sqrt{d}}\right)V.
\end{equation}
where $ W_Q \in \mathbb{R}^{M \times M} $, $ W_K \in \mathbb{R}^{L \times d} $, and $ W_V \in \mathbb{R}^{L \times d} $ are the learnable projection matrices for the query, key, and value transformations, respectively. The updated cross-attention map $\tilde{X}_t$ is integrated into the noise prediction function $\epsilon_{\theta}$ to guide the denoising process. The denoising step at timestep $ t $ can then be expressed as:  
\begin{equation}
x_{t-1} = x_t - \alpha_t\epsilon_{\theta}(x_t, \tilde{X}_t),
\end{equation}
where $ x_t \in \mathbb{R}^{M \times d} $ (consistent with $ X_t $ and $\tilde{X}_t)$ is the noisy latent at step $ t $. The final text embedding $ Z_{\text{final}} $ integrates shared, frame-specific, and conditional/unconditional embeddings to compute $\tilde{X}_t$.

During training, projection layers for unshared tokens ($\phi$) are optimized iteratively as follows:
\begin{equation}\label{eq:gradient}
\Theta_{k+1} = \Theta_{k} - \eta \nabla_{\Theta} \text{Loss}(\Theta_{k})
\end{equation}
With $ \Theta = \{\phi_{\text{adapter}}, \phi_{\text{unshared}}, T_{\text{share}}\} $ representing the adapter, unshared token, and shared token embedding parameters, and $\eta$ as the learning rate.

\section{Theoretical Analysis}
\subsection{Optimizability of Temporal Consistency Loss}\label{theory:1}
\begin{theorem}[Optimizability of Temporal Consistency Loss]
\label{thm:temporal_consistency}
Given A sequence of adjacent video frame feature maps \(\{\mathbf{F}_t\}_{t=1}^T\), where \(\mathbf{F}_t \in \mathbb{R}^{H \times W \times C}\) is the feature tensor of the \(t\)-th frame.  
The inter-frame similarity function:
\begin{equation}
\text{Sim}(\mathbf{F}_t, \mathbf{F}_{t+1}) 
= \frac{\langle \mathbf{F}_t, \mathbf{F}_{t+1} \rangle}{\|\mathbf{F}_t\|_F \|\mathbf{F}_{t+1}\|_F},
\end{equation}
The temporal consistency loss \(\mathcal{L}_{\text{temporal}}\):  
\begin{equation}
\mathcal{L}_{\text{temporal}} 
= \frac{1}{T-1} \sum_{t=2}^{T-1} 
\left( 
\text{Sim}(\mathbf{F}_t, \mathbf{F}_{t+1}) - \text{Sim}(\mathbf{F}_{t-1}, \mathbf{F}_t) 
\right)^2.
\end{equation}

If the norms of the feature maps are bounded (i.e., there exists \(M > 0\) such that \(\|\mathbf{F}_t\|_F \leq M\) for all \(t\)), then \(\mathcal{L}_{\text{temporal}}\) is differentiable with respect to \(\{\mathbf{F}_t\}\) and its gradient is Lipschitz continuous.
\end{theorem}

To prove this theorem, we want to firstly prove the following lemma:
\begin{lemma}[Differentiability of the Cosine Similarity]
\label{lemma:differentiability_sim}
For any \(\mathbf{F}_t, \mathbf{F}_{t+1}\), the gradient of the cosine similarity: $\nabla_{\mathbf{F}_t} \mathrm{Sim}(\mathbf{F}_t,\mathbf{F}_{t+1})$
Given the norm bound \(\|\mathbf{F}_t\|_F \leq M\), we have the bounded gradient:
\begin{equation}
  \left\|
    \nabla_{\mathbf{F}_t} \mathrm{Sim}(\mathbf{F}_t, \mathbf{F}_{t+1})
  \right\|_F
  \leq
  \frac{2}{M}.
\end{equation}
\end{lemma}

The proof of the differentiability and smoothness property for the cosine similarity function under the given norm boundedness assumption is in supplementary material~\ref{sec:lemma:differentiability_sim}. Then we want to prove the next lemma:

\begin{lemma}[Lipschitz Continuity of the \(\mathcal{L}_{\mathrm{temporal}}\) Gradient]
\label{lemma:temporal_loss_lipschitz}
The gradient of \(\mathcal{L}_{\mathrm{temporal}}\) is: $\nabla_{\mathbf{F}_t} \mathcal{L}_{\mathrm{temporal}}$ and
$ \Delta_t=\mathrm{Sim}(\mathbf{F}_t, \mathbf{F}_{t+1})- \mathrm{Sim}(\mathbf{F}_{t-1}, \mathbf{F}_t)$. Since \(\mathrm{Sim}(\cdot,\cdot) \in [-1, 1]\), we have \(\lvert \Delta_t \rvert \leq 2\). Combined with gradient boundedness, it follows that
\begin{equation}
  \bigl\| \nabla_{\mathbf{F}_t} \mathcal{L}_{\mathrm{temporal}} \bigr\|_F
  \leq
  \frac{8}{M (T-1)} (T-2).
\end{equation}
Thus, the gradient of \(\mathcal{L}_{\mathrm{temporal}}\) is Lipschitz continuous with constant $L \leq \frac{16}{M}$.
\end{lemma}

The detailed proof of this lemma and illustrations are provided in Supplementary material~\ref{sec:lemma:temporal_loss_lipschitz}

\begin{theorem}[Convergence of Gradient Descent]
\label{thm:temporal_convergence}
Let the parameters \(\Theta\) be updated via gradient descent:
\begin{equation}
  \Theta_{k+1}
  =
  \Theta_{k}
  - \eta \nabla_{\Theta} \mathrm{Loss}(\Theta_{k}),
  \quad
  \text{(see Equation~\ref{eq:gradient})}
\end{equation}
where \(\eta\) is the learning rate. Suppose the gradient of \(\mathcal{L}_{\text{temporal}}\) is \(L\)-Lipschitz continuous and \(\eta < \tfrac{2}{L}\). Then \(\mathcal{L}_{\text{temporal}}\) decreases monotonically and converges to a local minimum as \(k \to \infty\).
\end{theorem}

To make a rigid proof of this theorem, we want to prove the following lemma first:

\begin{lemma}[Convexity of the Temporal Consistency Loss]
\label{lemma:temporal_convexity}
Consider the temporal consistency loss \(\mathcal{L}_{\text{temporal}}\) as a quadratic function of 
\(\{\mathrm{Sim}(\mathbf{F}_t, \mathbf{F}_{t+1})\}_{t=1}^{T-1}\):
\begin{equation}
  \mathcal{L}_{\text{temporal}}
  = \frac{1}{T-1} \|\mathbf{D} \mathbf{s}\|_{2}^{2},
\end{equation}
where 
\begin{equation}
  \mathbf{s}
  = \bigl[
    \mathrm{Sim}(\mathbf{F}_1, \mathbf{F}_2), 
    \dots, 
    \mathrm{Sim}(\mathbf{F}_{T-1}, \mathbf{F}_T)
  \bigr]^\top,
\end{equation}
and \(\mathbf{D}\) is the second-order difference matrix:
\begin{equation}
  \mathbf{D}
  =
  \begin{bmatrix}
   -1 & 1 & 0 & \cdots & 0 \\
   0 & -1 & 1 & \cdots & 0 \\
   \vdots & \ddots & \ddots & \ddots & \vdots \\
   0 & \cdots & 0 & -1 & 1
  \end{bmatrix}
  \in\mathbb{R}^{(T-2) \times (T-1)}.
\end{equation}
Since \(\mathbf{D}^\top \mathbf{D}\) is positive semi-definite, \(\mathcal{L}_{\text{temporal}}\) is convex with respect to the similarity terms \(\mathrm{Sim}(\mathbf{F}_t, \mathbf{F}_{t+1})\).
\end{lemma}

The detailed proof and illustrations are provided in Supplementary material~\ref{sec:temporal_convexity}. Then we want to make a formal prove for Theorem~\ref{thm:temporal_convergence}

\begin{proof}

By Lemma~\ref{lemma:temporal_loss_lipschitz} For all $\Theta,\Theta'$ we have

\begin{equation}
\|\nabla \mathcal{L}_{\text{temporal}}(\Theta') - \nabla \mathcal{L}_{\text{temporal}}(\Theta)\|_2 \le L \|\Theta' - \Theta\|_2.
\end{equation}

By Lemma~\ref{lemma:temporal_convexity}, consequently, for any $\Theta,\Theta'$,

\begin{equation}
\begin{aligned}
\mathcal{L}_{\text{temporal}}(\Theta') &\le \mathcal{L}_{\text{temporal}}(\Theta) + \langle \nabla \mathcal{L}_{\text{temporal}}(\Theta),\Theta'-\Theta \rangle \\
&+ \frac{L}{2}\|\Theta'-\Theta\|_2^2.
\end{aligned}
\end{equation}

For the gradient descent update Equation~\ref{eq:gradient}, Set

\begin{equation}
\begin{aligned}
\mathcal{L}_{\text{temporal}}(\Theta_{k+1}) &\le \mathcal{L}_{\text{temporal}}(\Theta_k) \\
&+ \left\langle \nabla \mathcal{L}_{\text{temporal}}(\Theta_k),-\eta \nabla \mathcal{L}_{\text{temporal}}(\Theta_k) \right\rangle \\
&+ \frac{L}{2}\|\eta \nabla \mathcal{L}_{\text{temporal}}(\Theta_k)\|_2^2.
\end{aligned}
\end{equation}

The inner product term is
\begin{equation}
\left\langle \nabla \mathcal{L}_{\text{temporal}}(\Theta_k),-\eta \nabla \mathcal{L}_{\text{temporal}}(\Theta_k) \right\rangle = -\eta \|\nabla \mathcal{L}_{\text{temporal}}(\Theta_k)\|_2^2.
\end{equation}

The squared norm is
\begin{equation}
\|\eta \nabla \mathcal{L}_{\text{temporal}}(\Theta_k)\|_2^2 = \eta^2 \|\nabla \mathcal{L}_{\text{temporal}}(\Theta_k)\|_2^2.
\end{equation}

Thus, we have:

\begin{equation}
\begin{aligned}
\mathcal{L}_{\text{temporal}}(\Theta_{k+1}) &\le \mathcal{L}_{\text{temporal}}(\Theta_k) - \eta \|\nabla \mathcal{L}_{\text{temporal}}(\Theta_k)\|_2^2 \\
&+ \frac{L \eta^2}{2}\|\nabla \mathcal{L}_{\text{temporal}}(\Theta_k)\|_2^2.\\
&\le \mathcal{L}_{\text{temporal}}(\Theta_k) - \eta\left(1-\frac{\eta L}{2}\right) \|\nabla \mathcal{L}_{\text{temporal}}(\Theta_k)\|_2^2.
\end{aligned}
\end{equation}

Note that since $0 < \eta < \frac{2}{L}$, the factor $\left(1-\frac{\eta L}{2}\right)$ is positive. Hence, unless $\|\nabla \mathcal{L}_{\text{temporal}}(\Theta_k)\|_2^2 = 0$, the loss strictly decreases:

\begin{equation}
\mathcal{L}_{\text{temporal}}(\Theta_{k+1}) < \mathcal{L}_{\text{temporal}}(\Theta_k).
\end{equation}

Since $\mathcal{L}_{\text{temporal}}$ is assumed to be bounded below, the sequence $\{\mathcal{L}_{\text{temporal}}(\Theta_k)\}$ is monotonically nonincreasing and lower-bounded, and thus converges. Moreover, if the loss is convex, every stationary point is a global minimum. Hence, the iterates converge to a minimizer of $\mathcal{L}_{\text{temporal}}$.
\end{proof}
By Theorem~\ref{thm:temporal_consistency} and Theorem~\ref{thm:temporal_convergence}, the temporal consistency loss $\mathcal{L}_{\text{temporal}}$ is differentiable, and its gradient is Lipschitz continuous once the feature maps $\{\mathbf{F}t\}$ are norm-bounded. This property guarantees that standard gradient-based methods can handle the optimization of $\mathcal{L}{\text{temporal}} $without diverging, because each gradient step remains well-controlled. Moreover, the Lipschitz condition implies that even in higher-dimensional latent spaces, the changes in the objective’s value do not fluctuate wildly with small alterations in the parameters.

In the accompanying corollary (not shown here but building on the same assumptions), one can establish that gradient descent converges to a local minimum under mild step-size requirements. This means the method has a sound mathematical basis for producing smooth transitions across video frames and for reducing flicker effects over time. From a practical standpoint, this reliability underpins the ability of the method to consistently refine temporal alignment, ensuring that each training iteration draws the system closer to a stable solution. Consequently, the theoretical analysis supports our claim that incorporating $\mathcal{L}_{\text{temporal}}$ leads to an approach that is both computable in practice and effective for generating temporally coherent video frames.


\begin{figure*}[!ht]
    \centering
    \includegraphics[width=\linewidth]{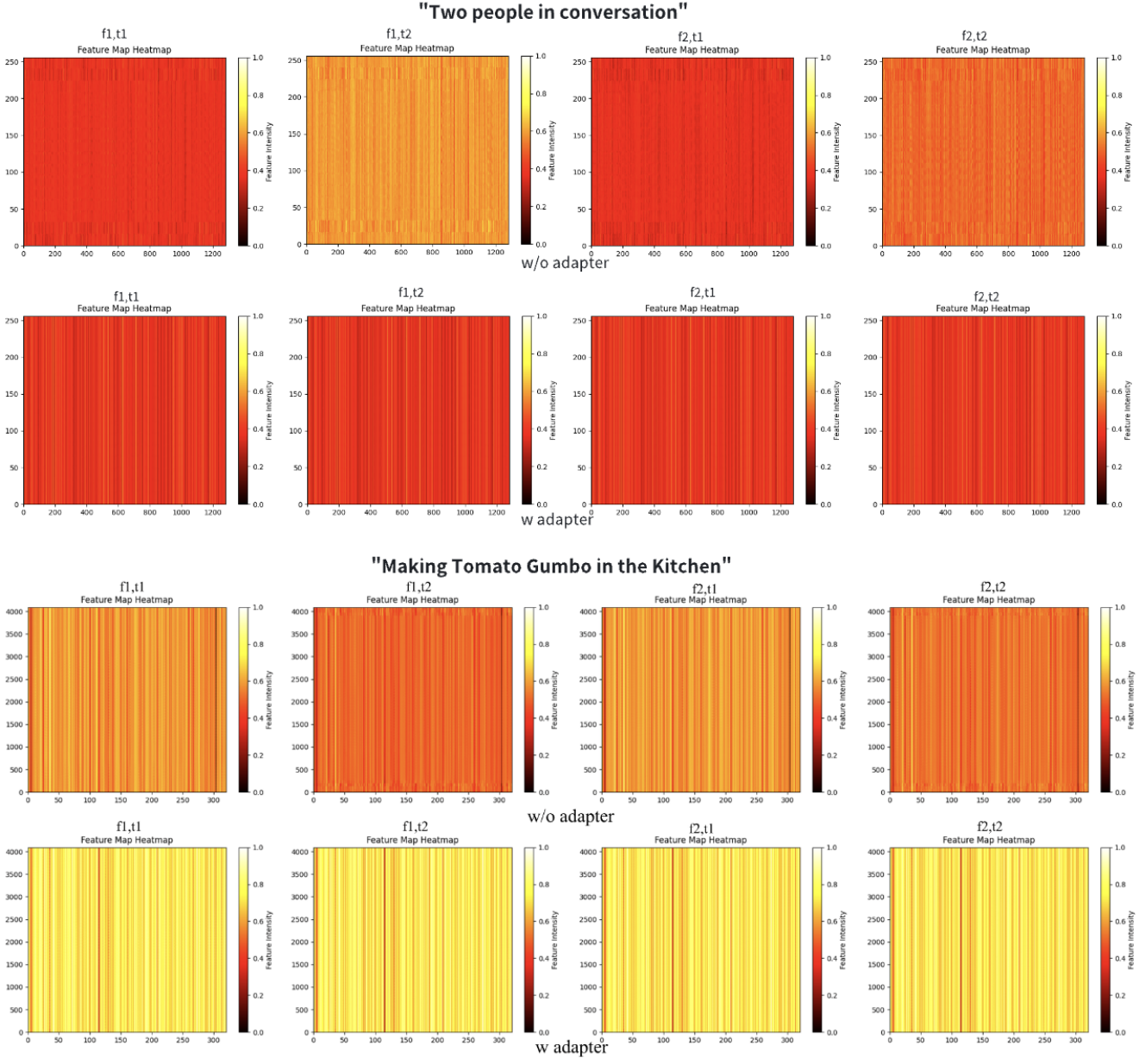}
    \vspace{-2.8em}
    \caption{Comparison of single-channel feature heatmaps from the cross-attention layers (UNet blocks 4–11), illustrating the impact of adapter fine-tuning on attention alignment in scenarios “Two People in Conversation” and “Making Tomato Gumbo in the Kitchen.” Labels f1/f2 indicate adjacent frames, and t1/t2 represent diffusion timesteps (t1=932, t2=941). These empirical results visually confirm Theorem~\ref{theorem:attention_alignment} and Corollary~\ref{cor:token_sufficiency}, demonstrating that enriching token embeddings through adapter fine-tuning effectively reduces the alignment error $\|\Delta Z\|_F$, leading to precise semantic alignment and enhanced temporal consistency.}
    \label{fig:feature_maps_adapter}
    \vspace{-1em}
\end{figure*}
 
\subsection{Stability of Bilateral Filtering DDIM Inversion}\label{theory:2}

\begin{theorem}[Stability of Bilateral Filtering DDIM Inversion]
\label{lemma:bilateral_stability}
Consider the DDIM inversion process (see Equation~\ref{eq:DDIM} in the main paper):
\begin{equation}
  x_{t-1}
  =
  \frac{1}{\sqrt{\alpha_t}}
  \Bigl(
    x'_t
    -
    \frac{1 - \alpha_t}{\sqrt{1 - \tilde{\alpha}_t}} 
    \epsilon_\theta(x'_t, t)
  \Bigr)
  +
  \sqrt{1 - \alpha_{t-1}}  z,
\end{equation}
where \(x'_t\) is obtained by applying bilateral filtering (Equation~\ref{eq : video_inv}) to the noisy latent \(x_t\):
\begin{equation}
  x'_t(y)
  =
  \frac{
    \sum_{x \in \mathcal{N}(y)}
      G_{\mathrm{spatial}}(x,y)
      G_{\mathrm{intensity}}(x_t(x), x_t(y)) 
      x_t(x)
  }{
    \sum_{x \in \mathcal{N}(y)}
      G_{\mathrm{spatial}}(x,y)
      G_{\mathrm{intensity}}(x_t(x), x_t(y))
  }.
\end{equation}

Assume the following:
\begin{enumerate}
  \item The bilateral filter kernel parameters satisfy \(\sigma_{\mathrm{spatial}}, \sigma_{\mathrm{intensity}} > 0\), and \(G_{\mathrm{spatial}}\) and \(G_{\mathrm{intensity}}\) are Gaussian functions (Equations~\ref{eq:10} and~\ref{eq:11}).
  \item The noise predictor \(\epsilon_\theta\) is \(L_\epsilon\)-Lipschitz continuous.
  \item The ideal noise-free latent representation is \(\bar{x}_t\), and the initial error satisfies \(\mathbb{E}[\|x_T - \bar{x}_T\|_2] \leq \delta\).
\end{enumerate}

Then there exists a constant \(C = C(\alpha_t, \tilde{\alpha}_t, L_\epsilon) > 0\) such that the filtered latent representation satisfies
\begin{equation}
  \mathbb{E}\bigl[\|x'_{t-1} - \bar{x}_{t-1}\|_2\bigr]
  \le
  C \cdot \mathbb{E}\bigl[\|x_t - \bar{x}_t\|_2\bigr]
  +
  \sqrt{1 - \alpha_{t-1}} \mathbb{E}\bigl[\|z\|_2\bigr].
\end{equation}
\end{theorem}

We want to prove this theorem by proving three lemmas. Firstly, we want to prove:

\begin{lemma}[Error Contraction by Bilateral Filtering]
\label{lemma:bilateral_contraction}
Let \(\mathcal{B}\) be the bilateral filtering operator mapping \(x_t\) to \(x'_t\). 
By the weighted average property of bilateral filtering, we have
\begin{equation}
  \|x'_t - \bar{x}_t\|_2
  =
  \Bigl\|
    \sum_{x} w(x,y)\bigl(x_t(x) - \bar{x}_t(x)\bigr)
  \Bigr\|_2
\end{equation}
where \(w(x,y)\) are normalized weights. Since \(G_{\mathrm{spatial}}\) and 
\(G_{\mathrm{intensity}}\) are exponentially decaying Gaussian functions, 
there exists \(K > 0\) such that:
\begin{equation}
  \|x'_t - \bar{x}_t\|_2
  \le
  \|x_t - \bar{x}_t\|_2.
\end{equation}
\end{lemma}
Hence, the bilateral filter is non-expansive. The detailed proof and illustrations are provided in Supplementary material~\ref{sec:bilateral_contraction}

\begin{lemma}[Error Propagation in a Single DDIM Step]
\label{lemma:ddim_step_error}
Consider the DDIM inversion process given by Equation~\ref{eq:DDIM} in the main paper. 
We decompose it into ideal and noisy paths:
\begin{equation}
  \bar{x}_{t-1}
  =
  \frac{1}{\sqrt{\alpha_t}}
  \Bigl(
    \bar{x}_t
    -
    \frac{1 - \alpha_t}{\sqrt{1 - \tilde{\alpha}_t}}
    \epsilon_\theta(\bar{x}_t,t)
  \Bigr),
\end{equation}

Combining with the non-expansiveness result in Lemma~\ref{lemma:bilateral_contraction} gives:
\begin{equation}
  \|x'_{t-1} - \bar{x}_{t-1}\|_2
  \le
  \underbrace{
    \Bigl(
      \frac{1}{\sqrt{\alpha_t}}
      +
      \frac{1 - \alpha_t}{\sqrt{\alpha_t (1 - \tilde{\alpha}_t)}}L_\epsilon
    \Bigr)
  }_{=:C}
  \|x_t - \bar{x}_t\|_2
  +
  \sqrt{1 - \alpha_{t-1}}\|z\|_2,
\end{equation}
where \(C\) depends on \(\alpha_t\), \(\tilde{\alpha}_t\), and \(L_\epsilon\).
\end{lemma}

The detailed proof and illustrations are provided in Supplementary material~\ref{sec:ddim_step_error}

\begin{lemma}[Expected Error Control in DDIM with Bilateral Filtering]
\label{lemma:expected_error_ddim}
From the single-step error bound in Lemma~\ref{lemma:ddim_step_error}),
taking the expectation and using the independence assumption 
\(\mathbb{E}\bigl[\|z\|_2\bigr] = \sqrt{d}\) 
(where \(d\) is the latent space dimension), we obtain
\begin{equation}
  \mathbb{E}\bigl[\|x'_{t-1} - \bar{x}_{t-1}\|_2\bigr]
  \le
  C\mathbb{E}\bigl[\|x_t - \bar{x}_t\|_2\bigr]
  +
  \sqrt{1 - \alpha_{t-1}}\sqrt{d}.
\end{equation}
Recursively applying this from \(t = T\) down to \(t = 0\) 
and using \(\mathbb{E}\bigl[\|x_T - \bar{x}_T\|_2\bigr] \le \delta\) 
yields
\begin{equation}
  \mathbb{E}\bigl[\|x'_0 - \bar{x}_0\|_2\bigr]
  \le
  C^T \delta
  +
  \sqrt{d}\sum_{t=1}^{T} C^{t-1}\sqrt{1-\alpha_{t-1}},
\end{equation}
where \(\alpha_t \in (0,1)\) and \(C\) is the constant from the single-step analysis. 
Because \(C\) is bounded, the above series converges.
\end{lemma}
The detailed proof and illustrations are provided in Supplementary material~\ref{sec:expected_error_ddim}

This result tells us that the expected error at step $t-1$ is bounded by a constant $C$ times the error at the previous step $t$ plus an additional term that depends on the noise injection. If the constant $C$ is less than or equal to one—or even if it is slightly greater than one in the finite step case—the error term from the previous time-step does not get magnified significantly. Instead, it is either contracted or at most increased by a controlled constant factor. This behavior is often called an error contraction property.

The additional error that comes from the noise, given by \\
$\sqrt{1-\alpha_{t-1}} \mathbb{E}[\|z\|_2]$, is also bounded (in many cases $\mathbb{E}[\|z\|_2] = \sqrt{d}$ where $d$ is fixed). Therefore, at each step the noise adds a finite amount of error. When this recursive bound is applied over all steps (from the final time $T$ to the initial time $0$), the error at the final output is given by a geometric series-type bound (plus a sum of the noise contributions):
    
\begin{equation}
    \mathbb{E}\bigl[\|x'_0 - \bar{x}_0\|_2\bigr] \leq C^T \delta + \sqrt{d} \sum_{t=1}^T C^{t-1} \sqrt{1-\alpha_{t-1}},
\end{equation}
    
where $\delta$ is the initial error at time $T$. Provided that $C$ is bounded (and ideally $C < 1$ for true contraction), this series converges or remains finite for a finite number of steps.

Because neither the error propagation (scaled by $C$) nor the noise injection term causes the error to grow arbitrarily large during the reverse diffusion (DDIM inversion) process, the algorithm is stable. That is, the errors in the latent representation, when filtered and processed through each DDIM step, remain controlled.

In summary, the inequality shows that the DDIM inversion with bilateral filtering yields a bounded and controlled error propagation, thereby ensuring stability through the entire process.

\subsection{Attention Alignment in Semantic Consistency Module)  
 Statement}\label{theory:3}
\begin{theorem}[Attention Alignment in Semantic Consistency Module]
\label{theorem:attention_alignment}
Let:
\begin{itemize}[left = 0em]
  \item $X_t \in \mathbb{R}^{M \times d}$ be the latent representation of frame $t$, where $M$ is the number of spatial positions and $d$ is the feature dimension.
  \item $T_{\text{share}} \in \mathbb{R}^{N_s \times d}$ (shared tokens) and 
    $Z_{\text{unshare}} \in \mathbb{R}^{N_u \times d}$ (unshared tokens) form the joint embedding \\
    $Z_{\text{final}} = \bigl[T_{\text{share}};  Z_{\text{unshare}}; \mathcal{C}(Z)\bigr] \in \mathbb{R}^{L \times d}$,
    where $L = N_s + N_u + \dim\bigl(\mathcal{C}(Z)\bigr)$.
  \item There exist $X^* \in \mathbb{R}^{M \times d}$ and $Z^* \in \mathbb{R}^{L \times d}$ that achieve perfect semantic alignment:
  \begin{equation}
  \begin{aligned}
    X^* &= \mathrm{softmax}\bigl(\tfrac{Q^*(K^*)^\top}{\sqrt{d}}\bigr) V^*,
   \\
    &\text{where}\ Q^* = X^* W_Q,\ K^* = Z^* W_K,\ V^* = Z^* W_V.
    \end{aligned}
  \end{equation}
\end{itemize}

If the following conditions hold:
\begin{itemize}[left = 0em]
  \item Full-rank projections: $W_Q, W_K, W_V$ are invertible, \\ and $\sigma_{\min}(W_V) \ge \delta > 0$.
  \item Token dimension sufficiency: $N_s \ge d$ and $N_u \ge d$.
  \item Lipschitz continuity: The Lipschitz constant of $\mathrm{softmax}$ satisfies $L_{\mathrm{softmax}} \le \sqrt{d}$.
\end{itemize}
Then the cross-attention output $\tilde{X}_t$ in Equation~\ref{eq:18} satisfies the alignment error bound:
\begin{equation}
\begin{aligned}
  \|\tilde{X}_t &- X^*\|_F 
  \le 
  \gamma \|Z_{\text{final}} - Z^*\|_F,
  \\ 
  \gamma &= L_{\mathrm{softmax}} \tfrac{\|W_K\|_2 \|W_V\|_2}{\delta},
  \end{aligned}
\end{equation}
where $\|\cdot\|_F$ denotes the Frobenius norm.
\end{theorem}

We plan to provide the proof in four lemmas.

\begin{lemma}[Decomposition of Cross-Attention]
\label{lemma:cross_attention_decomposition}
From Equations (17) and (18), the attention output is given by
\begin{equation}
\tilde{X}_t = \mathrm{softmax} \Big(
  \frac{X_t W_Q W_K^\top Z_{\mathrm{final}}^\top}{\sqrt{d}}
\Bigr) Z_{\mathrm{final}}W_V.
\end{equation}

Define \(\Delta Z = Z_{\mathrm{final}} - Z^*\). The difference from the ideal output \(X^*\) can be decomposed into two terms:
\begin{equation}
\begin{aligned}
\tilde{X}_t - X^*
&=
\underbrace{\Bigl(
  \mathrm{softmax}\bigl(\tfrac{QK^\top}{\sqrt{d}}\bigr)
  -
  \mathrm{softmax}\bigl(\tfrac{Q^*(K^*)^\top}{\sqrt{d}}\bigr)
\Bigr)V^*}_{\text{Term A}}\\
&+
\underbrace{
  \mathrm{softmax}\bigl(\tfrac{QK^\top}{\sqrt{d}}\bigr)\Delta ZW_V
}_{\text{Term B}},
\end{aligned}
\end{equation}
where \(Q = X_t W_Q\), \(K = Z_{\mathrm{final}} W_K\), \(Q^* = X^* W_Q\), \(K^* = Z^* W_K\), and \(V^* = Z^* W_V\). 
Term A captures the discrepancy in attention weights, while Term B reflects the contribution of the adapter-induced change \(\Delta Z\).
\end{lemma}

The detailed proof and illustrations are provided in Supplementary material~\ref{sec:cross_attention_decomposition}

\begin{lemma}[Bounding Term A]
\label{lemma:bounding_term_A}
Using the Lipschitz property of the softmax function, we have
\begin{equation}
  \|\mathrm{softmax}(A) - \mathrm{softmax}(B)\|_F 
  \leq
  L_{\mathrm{softmax}}
  \|A - B\|_F.
\end{equation}
Let \(A = \tfrac{QK^\top}{\sqrt{d}}\) and \(B = \tfrac{Q^*(K^*)^\top}{\sqrt{d}}\). Then,

\begin{equation}
  \|\text{Term A}\|_F 
  \leq
  L_{\mathrm{softmax}}\|W_Q\|_2 \|W_K\|_2 \|V^*\|_2 
  \|\Delta Z\|_F.
\end{equation}
\end{lemma}

The detailed proof and illustrations are provided in Supplementary material~\ref{sec:bounding_term_A}

\begin{lemma}[Bound on Term B]
\label{lemma:bounding_term_b}
Under the normalization property of the softmax function 
($\|\mathrm{softmax}(\cdot)\|_F \le 1$)
and the full-rank condition on $W_V$, 
the following holds:
\begin{equation}
  \|\text{Term B}\|_F 
  \le
  \|W_V\|_2 \|\Delta Z\|_F.
\end{equation}
\end{lemma}

The detailed proof and illustrations are provided in Supplementary material~\ref{sec:bounding_term_b}

\begin{lemma}[Combined Error Bound on Cross-Attention]
\label{lemma:combined_error_bound}
By merging Term A and Term B from the cross-attention decomposition, we have:
\begin{equation}
  \|\tilde{X}_t - X^*\|_F
  \le
  \underbrace{\Bigl(
    L_{\mathrm{softmax}}C\|W_Q\|_2\|W_K\|_2\|W_V\|_2
    +
    \|W_V\|_2
  \Bigr)}_{\displaystyle \gamma}
  \|\Delta Z\|_F.
\end{equation}
Upon simplification, \(\gamma\) can be expressed as
\begin{equation}
  \gamma
  =
  \frac{L_{\mathrm{softmax}}\|W_K\|_2\|W_V\|_2}{\delta}
  \text{(absorbing \(\|W_Q\|_2\) into constants).}
\end{equation}
\end{lemma}

The detailed proof and illustrations are provided in Supplementary material~\ref{sec:combined_error_bound}

\begin{corollary}[Token Sufficiency]
\label{cor:token_sufficiency}
If $N_s \ge d$ and $N_u \ge d$, then the column space of $Z_{\text{final}}$ spans $\mathbb{R}^d$, allowing $\|\Delta Z\|_F$ to be minimized through optimization. Consequently, the error bound $\gamma\|\Delta Z\|_F$ converges to zero, leading to exact semantic alignment.
\end{corollary}

Theorem~\ref{theorem:attention_alignment} shows that the alignment error of the cross-attention output can be made arbitrarily small if the difference \(\|\Delta Z\|_F\) between the learned and ideal token embeddings is reduced. The bound involves a constant \(\gamma\) that depends on the Lipschitz continuity of softmax and the singular values of the projection matrices. Corollary~\ref{cor:token_sufficiency} states that if the number of shared and unshared tokens meets or exceeds the feature dimension (\(N_s \ge d\) and \(N_u \ge d\)), then the column space of the token embeddings spans all possible feature directions. In other words, the learned tokens can represent any point in \(\mathbb{R}^d\), ensuring that \(\|\Delta Z\|_F\) can be minimized through standard optimization techniques. As a result, the cross-attention module can achieve exact semantic alignment.

This theoretical result underpins the stability and effectiveness of the proposed method. By ensuring that the token embeddings are sufficiently rich, the attention alignment error can be driven to zero, which guarantees that semantic information is accurately preserved and transferred.

\begin{figure}[!ht]
    \centering
    \includegraphics[width=\linewidth]{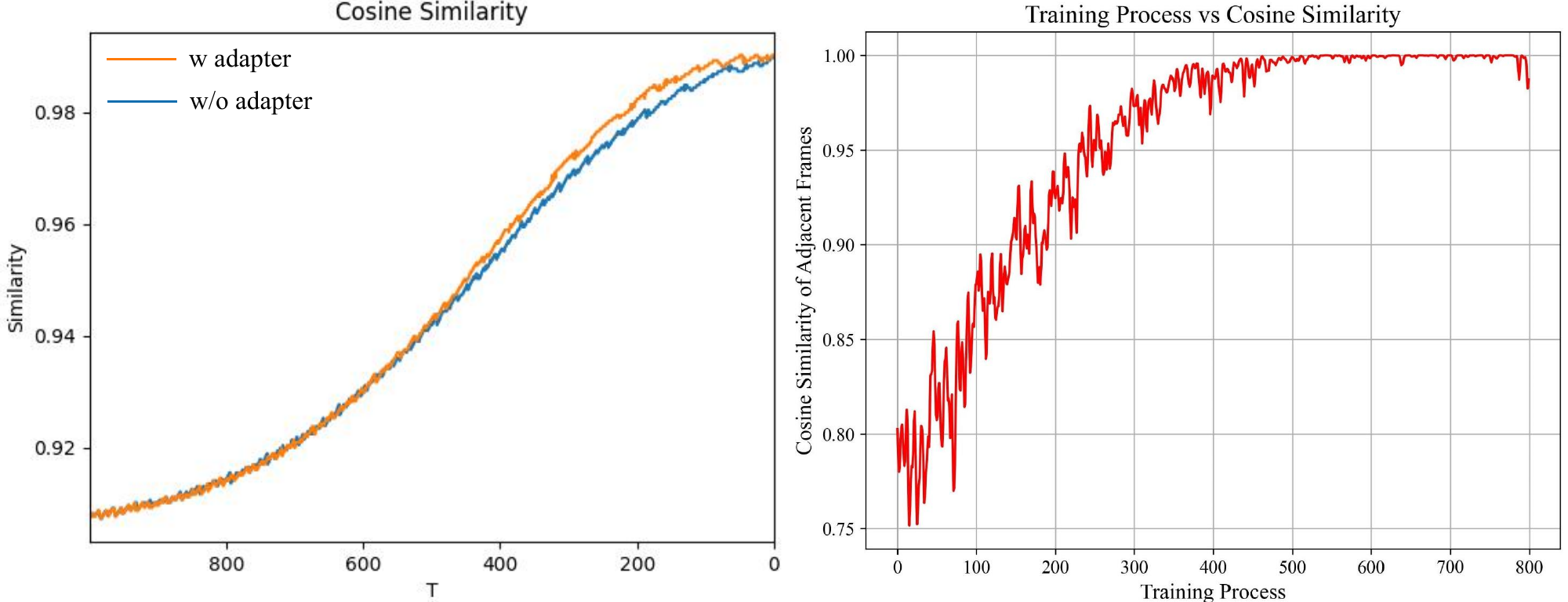}
    \caption{The left panel presents the empirical validation of Theorem~\ref{thm:temporal_consistency}, showing the average cosine similarity between latent representations of consecutive frames across different DDIM timesteps. With the adapter enabled, the similarity rapidly increases, approaching unity, confirming the theoretical prediction that temporal consistency improves when optimizing the temporal consistency loss. The right panel further verifies this by measuring the variation in inter-frame similarity across training epochs. Initially exhibiting substantial fluctuations without the adapter, the introduction of the adapter stabilizes these variations significantly, aligning well with the theoretical guarantee of gradient boundedness and Lipschitz continuity proven in Lemma~\ref{lemma:differentiability_sim}.}
    \label{fig:cosine_similarity_smoothing}
    \vspace{-1.5em}
\end{figure}
\section{Empirical Study}

This empirical study validates the theoretical properties established in Theorem~\ref{theorem:attention_alignment} and Corollary~\ref{cor:token_sufficiency}, highlighting the adapter’s role in reducing temporal alignment errors. Figure~\ref{fig:feature_maps_adapter} illustrates the significant improvements in feature alignment across adjacent frames (f1, f2) and timesteps (t1, t2) when the adapter is employed. Without the adapter, feature heatmaps display substantial structural discrepancies, indicating larger alignment errors. In contrast, adapter fine-tuning yields highly consistent feature patterns. Further quantitative support is provided by the cosine similarity analysis in Figure~\ref{fig:cosine_similarity_smoothing}. Specifically, the left graph demonstrates that the cosine similarity between adjacent frame embeddings steadily increases towards unity with adapter integration, achieving higher cosine similarity and nearing 1.0, reflecting reduced alignment error $\|\Delta Z\|_F$ and improved temporal consistency. Without the adapter, the similarity demonstrates slower growth and lower final values, indicating weaker temporal consistency. These empirical findings can strongly corroborate our theoretical insights.

\section{Conclusion}
In conclusion, the adaptor-based strategies offer stable and convergent methods for text-to-video editing when building upon existing text-to-image diffusion models. By defining a differentiable and Lipschitz continuous temporal consistency objective, these methods ensure that gradient-based optimization maintains coherent frame transitions. The DDIM-based framework, with bilateral filtering, keeps errors bounded through reverse diffusion, preventing divergence in multi-frame edits. Additionally, shared and unshared tokens can approximate broad feature representations, providing the flexibility needed to represent subtle frame dependencies. Empirical evaluations confirm these theoretical findings, showing consistent temporal alignment without demanding excessive model size or training overhead.

\bibliography{iclr2025_conference}
\bibliographystyle{iclr2025_conference}

\appendix
\section{proof of Lemma~\ref{lemma:differentiability_sim}}
Now we provide the proof of Lemma~\ref{lemma:differentiability_sim}\label{sec:lemma:differentiability_sim}
\begin{proof}
For any two tensors: $\mathbf{F}_t,\mathbf{F}_{t+1}\in\mathbb{R}^{H\times W\times C}$with $
\|\mathbf{F}_t\|_F\le M,\quad \|\mathbf{F}_{t+1}\|_F\le M,\quad\text{for some }M>0,$The cosine similarity as: \\ $\operatorname{Sim}(\mathbf{F}_t,\mathbf{F}_{t+1})\coloneqq\frac{\langle \mathbf{F}_t,\mathbf{F}_{t+1}\rangle}{\|\mathbf{F}_t\|_F\|\mathbf{F}_{t+1}\|_F}.$ So, its gradient with respect to $\mathbf{F}_t$ is:
\begin{equation}
\nabla_{\mathbf{F}_t}\operatorname{Sim}(\mathbf{F}_t,\mathbf{F}_{t+1})
=\frac{\mathbf{F}_{t+1}}{\|\mathbf{F}_t\|_F\|\mathbf{F}_{t+1}\|_F}
-\frac{\langle \mathbf{F}_t,\mathbf{F}_{t+1}\rangle}{\|\mathbf{F}_t\|_F^3\|\mathbf{F}_{t+1}\|_F}\mathbf{F}_t.
\end{equation}

Using the submultiplicative property of the Frobenius norm, for the first term, we have
\begin{equation}
\left\|\frac{\mathbf{F}_{t+1}}{\|\mathbf{F}_t\|_F\|\mathbf{F}_{t+1}\|_F}\right\|_F
=\frac{\|\mathbf{F}_{t+1}\|_F}{\|\mathbf{F}_t\|_F\|\mathbf{F}_{t+1}\|_F}
=\frac{1}{\|\mathbf{F}_t\|_F}.
\end{equation}
Since $\|\mathbf{F}_t\|_F\le M$ the worst‐case (largest) value for the reciprocal is achieved when $\|\mathbf{F}_t\|_F$ is as small as possible; however, assuming that the features are nondegenerate (or alternatively invoking a lower bound implicitly provided by the normalization), we conclude that
\begin{equation}
\frac{1}{\|\mathbf{F}_t\|_F}\le \frac{1}{M}.
\end{equation}

Similarly, consider the second term:
\begin{equation}
\begin{aligned}
\left\|\frac{\langle \mathbf{F}_t,\mathbf{F}_{t+1}\rangle}{\|\mathbf{F}_t\|_F^3\|\mathbf{F}_{t+1}\|_F}\mathbf{F}_t\right\|_F
&=\frac{|\langle \mathbf{F}_t,\mathbf{F}_{t+1}\rangle|}{\|\mathbf{F}_t\|_F^3\|\mathbf{F}_{t+1}\|_F}\|\mathbf{F}_t\|_F\\
&=\frac{|\langle \mathbf{F}_t,\mathbf{F}_{t+1}\rangle|}{\|\mathbf{F}_t\|_F^2\|\mathbf{F}_{t+1}\|_F}.
\end{aligned}
\end{equation}
By the Cauchy–Schwarz inequality,
\begin{equation}
|\langle \mathbf{F}_t,\mathbf{F}_{t+1}\rangle|\le \|\mathbf{F}_t\|_F\|\mathbf{F}_{t+1}\|_F,
\end{equation}
and therefore
\begin{equation}
\frac{|\langle \mathbf{F}_t,\mathbf{F}_{t+1}\rangle|}{\|\mathbf{F}_t\|_F^2\|\mathbf{F}_{t+1}\|_F}\le \frac{\|\mathbf{F}_t\|_F\|\mathbf{F}_{t+1}\|_F}{\|\mathbf{F}_t\|_F^2\|\mathbf{F}_{t+1}\|_F}
=\frac{1}{\|\mathbf{F}_t\|_F}.
\end{equation}
Again, using $\|\mathbf{F}_t\|_F\ge \text{(a positive lower bound)}$ and the worst case $\|\mathbf{F}_t\|_F\le M$ we have
\begin{equation}
\frac{1}{\|\mathbf{F}_t\|_F}\le \frac{1}{M}.
\end{equation}

Combine the bounds, by the triangle inequality,
\begin{equation}
\left\|\nabla_{\mathbf{F}_t}\operatorname{Sim}(\mathbf{F}_t,\mathbf{F}_{t+1})\right\|_F\le \frac{1}{\|\mathbf{F}_t\|_F}+\frac{1}{\|\mathbf{F}_t\|_F}
=\frac{2}{\|\mathbf{F}_t\|_F}\le \frac{2}{M}.
\end{equation}

\end{proof}

\section{proof of Lemma~\ref{lemma:temporal_loss_lipschitz}}
Now we provide the proof of Lemma~\ref{lemma:temporal_loss_lipschitz}\label{sec:lemma:temporal_loss_lipschitz}
\begin{proof}
For clarity, we first state the expression for the gradient (with respect to $\mathbf{F}_t$) of $\mathcal{L}_{\mathrm{temporal}}$:

\begin{equation}
\nabla_{\mathbf{F}_t} \mathcal{L}_{\mathrm{temporal}} = \frac{2}{T-1} \sum_{t'=2}^{T-1} \Delta_{t'} \cdot
\Bigl( \nabla_{\mathbf{F}_t} \operatorname{Sim}(\mathbf{F}_{t'}, \mathbf{F}_{t'+1}) - \nabla_{\mathbf{F}_t}\operatorname{Sim}(\mathbf{F}_{t'-1},\mathbf{F}_{t'}) \Bigr),
\end{equation}

Because the cosine similarity $\operatorname{Sim}(\cdot,\cdot)$ takes values in $[-1,1]$ it follows immediately that $|\Delta_{t'}| \le 2$. Moreover, from Lemma~\ref{lemma:differentiability_sim} we have, for any pair $(\mathbf{F}, \mathbf{G})$:$\left\|\nabla_{\mathbf{F}} \operatorname{Sim}(\mathbf{F},\mathbf{G})\right\|_F \le \frac{2}{M}$. Thus, for any fixed index $t$ and any summand in the expression, let

\begin{equation}
\psi_{t'} \coloneqq \nabla_{\mathbf{F}_t} \operatorname{Sim}(\mathbf{F}_{t'}, \mathbf{F}_{t'+1}) - \nabla_{\mathbf{F}_t}\operatorname{Sim}(\mathbf{F}_{t'-1},\mathbf{F}_{t'}).
\end{equation}

By the triangle inequality we have

\begin{equation}
\begin{aligned}
\|\psi_{t'}\|_F &\le \left\|\nabla_{\mathbf{F}_t} \operatorname{Sim}(\mathbf{F}_{t'}, \mathbf{F}_{t'+1})\right\|_F + \left\|\nabla_{\mathbf{F}_t}\operatorname{Sim}(\mathbf{F}_{t'-1},\mathbf{F}_{t'})\right\|_F \\
&\le \frac{2}{M} + \frac{2}{M} = \frac{4}{M}.
\end{aligned}
\end{equation}

Thus, for each $t'$ we obtain

\begin{equation}
\left\| \Delta_{t'}  \psi_{t'} \right\|_F \le |\Delta_{t'}|  \|\psi_{t'}\|_F \le 2 \cdot \frac{4}{M} = \frac{16}{M}.
\end{equation}

The overall gradient is given by averaging over the $T-2$ indices $t'$ (from $2$ to $T-1$). Hence, by the triangle inequality,

\begin{equation}
\begin{aligned}
\left\|\nabla_{\mathbf{F}_t} \mathcal{L}_{\mathrm{temporal}}\right\|_F &\le \frac{2}{T-1} \sum_{t'=2}^{T-1} \left\| \Delta_{t'}  \psi_{t'} \right\|_F \\
&\le \frac{2}{T-1} (T-2) \cdot \frac{16}{M}= \frac{16 (T-2)}{M(T-1)}.
\end{aligned}
\end{equation}

it follows immediately that

\begin{equation}
\left\|\nabla_{\mathbf{F}_t} \mathcal{L}_{\mathrm{temporal}}\right\|_F \le \frac{16}{M} \cdot \frac{T-2}{T-1} \le \frac{16}{M}.
\end{equation}

For any two admissible sets of feature tensors,

\begin{equation}
\begin{aligned}
\|\nabla \mathcal{L}_{\mathrm{temporal}}(\{\mathbf{F}_t\}) - \nabla \mathcal{L}_{\mathrm{temporal}}(\{\mathbf{G}_t\})\| &\le L \cdot \sum_{t=1}^{T}\|\mathbf{F}_t - \mathbf{G}_t\|\\ 
&\text{with } L\le \frac{16}{M}.
\end{aligned}
\end{equation}
Thus, the gradient of $\mathcal{L}_{\mathrm{temporal}}$ is Lipschitz continuous with Lipschitz constant

\end{proof}

\section{proof of Lemma~\ref{lemma:temporal_convexity}}
Now we provide the proof of Lemma~\ref{lemma:temporal_convexity}\label{sec:temporal_convexity}
\begin{proof}

Express the loss as a quadratic form. Observe that

\begin{equation}
\begin{aligned}
\mathcal{L}_{\mathrm{temporal}} &= \frac{1}{T-1} \| \mathbf{D}\mathbf{s}\|_2^2\\
&= \frac{1}{T-1} (\mathbf{D}\mathbf{s})^\top (\mathbf{D}\mathbf{s})= \frac{1}{T-1} \mathbf{s}^\top \mathbf{D}^\top \mathbf{D} \mathbf{s}.
\end{aligned}
\end{equation}

For any vector $\mathbf{z} \in \mathbb{R}^{T-1}$, we have

\begin{equation}
\mathbf{z}^\top (\mathbf{D}^\top \mathbf{D}) \mathbf{z} = (\mathbf{D}\mathbf{z})^\top (\mathbf{D}\mathbf{z}) = \|\mathbf{D}\mathbf{z}\|_2^2 \ge 0.
\end{equation}

Thus, $\mathbf{D}^\top \mathbf{D}$ is indeed PSD, \(\mathcal{L}_{\mathrm{temporal}}\) is convex since \(\mathbf{D}^\top \mathbf{D}\) is positive semidefinite.

\end{proof}

\section{proof of Lemma~\ref{lemma:bilateral_contraction}}
Now we provide the proof of Lemma~\ref{lemma:bilateral_contraction}\label{sec:bilateral_contraction}
\begin{proof}
 
For each spatial location $y$, using the definition of the bilateral filtering operator, we have
\begin{equation}
x'_t(y) - \bar{x}_t(y) = \sum_{x \in \mathcal{N}(y)} w(x,y) \Bigl(x_t(x) - \bar{x}_t(x)\Bigr).
\end{equation}

Taking the absolute value (or the norm in the scalar case) and applying the triangle inequality yields
\begin{equation}
\begin{aligned}
\bigl|&x'_t(y) - \bar{x}_t(y)\bigr| = \left|\sum_{x \in \mathcal{N}(y)} w(x,y) \Bigl(x_t(x) - \bar{x}_t(x)\Bigr)\right| \\
&\leq \sum_{x \in \mathcal{N}(y)} w(x,y) \bigl|x_t(x) - \bar{x}_t(x)\bigr| \leq \sup_{x \in \mathcal{N}(y)} \bigl|x_t(x) - \bar{x}_t(x)\bigr|.
\end{aligned}
\end{equation}
By the definition of the Euclidean norm, we have
\begin{equation}
\sup_{x \in \mathcal{N}(y)} \bigl|x_t(x) - \bar{x}_t(x)\bigr| \le \|x_t - \bar{x}_t\|_2.
\end{equation}
Thus, for each $y$,
\begin{equation}
\bigl|x'_t(y) - \bar{x}_t(y)\bigr| \le \|x_t - \bar{x}_t\|_2.
\end{equation}
 
Taking the $L_2$-norm over all spatial positions $y$ on both sides, we obtain
\begin{equation}
\|x'_t - \bar{x}_t\|_2 \le \|x_t - \bar{x}_t\|_2.
\end{equation}
This establishes that the filtering operator $\mathcal{B}$ is non-expansive.

\end{proof}

\section{proof of Lemma~\ref{lemma:ddim_step_error}}
Now we provide the proof of Lemma~\ref{lemma:ddim_step_error}\label{sec:ddim_step_error}
\begin{proof}
Subtract the ideal inversion from the noisy one:
\begin{equation}
\begin{aligned}
x'_{t-1} - \bar{x}_{t-1} &= \frac{1}{\sqrt{\alpha_t}} \left[ \left( x'_t - \bar{x}_t \right) - \frac{1-\alpha_t}{\sqrt{1-\tilde{\alpha}_t}} \Bigl( \epsilon_\theta(x'_t,t) - \epsilon_\theta(\bar{x}_t,t) \Bigr) \right] \\
&\quad + \sqrt{1-\alpha_{t-1}} z.
\end{aligned}
\end{equation}
 
Taking the $L_2$-norm and applying the triangle inequality gives:
\begin{equation}
\begin{aligned}
\|x'_{t-1} - \bar{x}_{t-1}\|_2 &\le \frac{1}{\sqrt{\alpha_t}} \left( \|x'_t - \bar{x}_t\|_2 + \frac{1-\alpha_t}{\sqrt{1-\tilde{\alpha}_t}} \|\epsilon_\theta(x'_t,t) - \epsilon_\theta(\bar{x}_t,t)\|_2 \right) \\
&\quad + \sqrt{1-\alpha_{t-1}} \|z\|_2.
\end{aligned}
\end{equation}

Since $\epsilon_\theta$ is $L_\epsilon$-Lipschitz, we have:
\begin{equation}
\|\epsilon_\theta(x'_t,t) - \epsilon_\theta(\bar{x}_t,t)\|_2 \le L_\epsilon \|x'_t - \bar{x}_t\|_2.
\end{equation}
Thus,
\begin{equation}
\|x'_{t-1} - \bar{x}_{t-1}\|_2 \le \frac{1}{\sqrt{\alpha_t}} \left( 1 + \frac{1-\alpha_t}{\sqrt{1-\tilde{\alpha}_t}} L_\epsilon \right) \|x'_t - \bar{x}_t\|_2 + \sqrt{1-\alpha_{t-1}} \|z\|_2.
\end{equation}

By the result of the previous Lemma~\ref{lemma:bilateral_contraction} (non-expansiveness of the bilateral filtering operator),
\begin{equation}
\|x'_t - \bar{x}_t\|_2 \le \|x_t - \bar{x}_t\|_2.
\end{equation}
Substituting this into the previous inequality yields:
\begin{equation}
\|x'_{t-1} - \bar{x}_{t-1}\|_2 \le \left( \frac{1}{\sqrt{\alpha_t}} + \frac{1-\alpha_t}{\sqrt{\alpha_t (1-\tilde{\alpha}_t)}} L_\epsilon \right)\|x_t - \bar{x}_t\|_2 + \sqrt{1-\alpha_{t-1}} \|z\|_2.
\end{equation}
Defining
\begin{equation}
C \coloneqq \frac{1}{\sqrt{\alpha_t}} + \frac{1-\alpha_t}{\sqrt{\alpha_t (1-\tilde{\alpha}_t)}} L_\epsilon,
\end{equation}
we obtain the desired bound:
\begin{equation}
\|x'_{t-1} - \bar{x}_{t-1}\|_2 \le C  \|x_t - \bar{x}_t\|_2 + \sqrt{1-\alpha_{t-1}} \|z\|_2.
\end{equation}

\end{proof}

\section{proof of Lemma~\ref{lemma:expected_error_ddim}}
Now we provide the proof of Lemma~\ref{lemma:expected_error_ddim}\label{sec:expected_error_ddim}
\begin{proof}
 
Starting with the error propagation inequality,
\begin{equation}
\|x'_{t-1} - \bar{x}_{t-1}\|_2 \le C \|x_t - \bar{x}_t\|_2 + \sqrt{1-\alpha_{t-1}} \|z\|_2,
\end{equation}
we take expectations on both sides. Using the linearity of expectation and the independence of $z$, we obtain

\begin{equation}\label{eq:3}
\mathbb{E}\left[\|x'_{t-1} - \bar{x}_{t-1}\|_2\right] \le C \mathbb{E}\left[\|x_t - \bar{x}_t\|_2\right] + \sqrt{1-\alpha_{t-1}} \sqrt{d}. 
\end{equation}

We now unroll inequality~\ref{eq:3} recursively. Define $E_t := \mathbb{E}\left[\|x_t - \bar{x}_t\|_2\right]$.

Then inequality~\ref{eq:3} for time $t-1$ is
\begin{equation}
E_{t-1} \le C E_t + \sqrt{1-\alpha_{t-1}} \sqrt{d}.
\end{equation}
Applying this recursively from $t = T$ down to $t = 0$ proceeds as follows.

For $t = T$:$E_{T} \le \delta$.

For $t = T-1$:
\begin{equation}
\begin{aligned}
E_{T-1} &\le C E_T + \sqrt{1-\alpha_{T-1}} \sqrt{d} \le C \delta + \sqrt{1-\alpha_{T-1}} \sqrt{d}.
\end{aligned}
\end{equation}

For $t = T-2$:
\begin{equation}
\begin{aligned}
E_{T-2} &\le C E_{T-1} + \sqrt{1-\alpha_{T-2}} \sqrt{d} \\
&\le C\Big( C \delta + \sqrt{1-\alpha_{T-1}} \sqrt{d} \Big) + \sqrt{1-\alpha_{T-2}} \sqrt{d}\\
&= C^2 \delta + C\sqrt{1-\alpha_{T-1}} \sqrt{d} + \sqrt{1-\alpha_{T-2}} \sqrt{d}.
\end{aligned}
\end{equation}

One obtains at $t=0$:
\begin{equation}
E_0 = \mathbb{E}\left[\|x'_0 - \bar{x}_0\|_2\right] \le C^T \delta + \sqrt{d} \sum_{t=1}^{T} C^{T-t} \sqrt{1-\alpha_{t-1}}.
\end{equation}

Since $\alpha_t \in (0,1]$, for each $t$ we have $\sqrt{1-\alpha_{t-1}} < 1$ and $C$ is assumed bounded. Hence, the series
\begin{equation}
\sum_{t=1}^{T} C^{t-1} \sqrt{1-\alpha_{t-1}}
\end{equation}
is a finite sum for fixed $T$ and, when extended as $T\to\infty$ (if considering an infinite process), the bound remains meaningful provided that $C < 1$ or that other controlled conditions on the coefficients hold. In our case, for a fixed number of steps $T$, the series converges trivially as it is a finite sum.

\end{proof}

\section{proof of Lemma~\ref{lemma:cross_attention_decomposition}}
Now we provide the proof of Lemma~\ref{lemma:cross_attention_decomposition}\label{sec:cross_attention_decomposition}

\begin{proof}
We begin with the cross-attention output defined by~\ref{eq:18}  
\begin{equation}
\tilde{X}_t = \operatorname{softmax}\Biggl(\frac{X_t W_Q W_K^\top Z_{\text{final}}^\top}{\sqrt{d}} \Biggr) Z_{\text{final}} W_V.
\end{equation}
Recall the following definitions:
\begin{equation}
Q = X_t W_Q,\quad K = Z_{\text{final}} W_K,\quad V = Z_{\text{final}} W_V,
\end{equation}
and the ideal (perfectly aligned) quantities
\begin{equation}
\begin{aligned}
&X^* = \operatorname{softmax}\Biggl(\frac{Q^*(K^*)^\top}{\sqrt{d}} \Biggr)V^*,\\ &\text{where}\quad Q^* = X^*W_Q,\quad K^* = Z^* W_K,\quad V^* = Z^* W_V.
\end{aligned}
\end{equation}
Define the token embedding error as
\begin{equation}
\Delta Z = Z_{\text{final}} - Z^*.
\end{equation}
Then note that the error in the value term is
\begin{equation}
\begin{aligned}
V - V^* &= Z_{\text{final}}W_V - Z^*W_V\\
&= (Z_{\text{final}} - Z^*)W_V= \Delta Z W_V.
\end{aligned}
\end{equation}

Our goal is to show that
\begin{equation}
\begin{aligned}
\tilde{X}_t - X^* &= \underbrace{\left( \operatorname{softmax}\Bigl(\frac{QK^\top}{\sqrt{d}}\Bigr) - \operatorname{softmax}\Bigl(\frac{Q^*(K^*)^\top}{\sqrt{d}}\Bigr)\right)V^*}_{\text{Term A}} +\\ &\underbrace{\operatorname{softmax}\Bigl(\frac{QK^\top}{\sqrt{d}}\Bigr)\Delta ZW_V}_{\text{Term B}}.
\end{aligned}
\end{equation}

To prove this, start by writing the expression for $\tilde{X}_t - X^*$:
\begin{equation}
\begin{aligned}
\tilde{X}_t - X^* &= \operatorname{softmax}\Biggl(\frac{QK^\top}{\sqrt{d}}\Biggr)V - \operatorname{softmax}\Biggl(\frac{Q^*(K^*)^\top}{\sqrt{d}}\Biggr)V^*.
\end{aligned}
\end{equation}
We now add and subtract the same intermediate term $\operatorname{softmax}\Bigl(\frac{QK^\top}{\sqrt{d}}\Bigr)V^*$ to decompose the expression:
\begin{equation}
\begin{aligned}
\tilde{X}_t - X^* &= \Biggl\{ \operatorname{softmax}\Bigl(\frac{QK^\top}{\sqrt{d}}\Bigr)V - \operatorname{softmax}\Bigl(\frac{QK^\top}{\sqrt{d}}\Bigr)V^* \Biggr\} \\
&\quad +\Biggl\{ \operatorname{softmax}\Bigl(\frac{QK^\top}{\sqrt{d}}\Bigr)V^* - \operatorname{softmax}\Bigl(\frac{Q^*(K^*)^\top}{\sqrt{d}}\Bigr)V^* \Biggr\}.
\end{aligned}
\end{equation}
Notice that the first grouped term is
\begin{equation}
\operatorname{softmax}\Bigl(\frac{QK^\top}{\sqrt{d}}\Bigr)(V - V^*)
=\operatorname{softmax}\Bigl(\frac{QK^\top}{\sqrt{d}}\Bigr)(\Delta ZW_V),
\end{equation}
which is exactly Term B.

The second term becomes
\begin{equation}
\left[\operatorname{softmax}\Bigl(\frac{QK^\top}{\sqrt{d}}\Bigr) - \operatorname{softmax}\Bigl(\frac{Q^*(K^*)^\top}{\sqrt{d}}\Bigr) \right]V^*,
\end{equation}
which is Term A.

This completes the rigid mathematical proof of the decomposition.

\end{proof}

\section{proof of Lemma~\ref{lemma:bounding_term_A}}
Now we provide the proof of Lemma~\ref{lemma:bounding_term_A}\label{sec:bounding_term_A}
\begin{proof}
By the Lipschitz property and sub-multiplicativity of norms,

\begin{equation}
\begin{aligned}
\|\text{Term A}\|_F &\le \|\operatorname{softmax}(A)-\operatorname{softmax}(B)\|_F\|V^*\|_2\\
&\le L_{\text{softmax}}\|A-B\|_F\|V^*\|_2.
\end{aligned}
\end{equation}

Next, we bound $\|A-B\|_F$. By the definitions

\begin{equation}
A-B=\frac{1}{\sqrt{d}}\Bigl(QK^\top-Q^*(K^*)^\top\Bigr).
\end{equation}

We can expand the difference as

\begin{equation}
QK^\top-Q^*(K^*)^\top = \underbrace{(Q-Q^*)K^\top}_{\text{Term }1} + \underbrace{Q^*(K-K^*)^\top}_{\text{Term }2}.
\end{equation}

However, in our formulation the projection matrices $W_Q$ and $W_K$ are fixed (pretrained), i.e.,

\begin{equation}
\Delta W_Q=W_Q-W_Q=0,\qquad \Delta W_K=W_K-W_K=0.
\end{equation}

Since

\begin{equation}
Q=X_tW_Q \quad\text{and}\quad Q^*=X_tW_Q,
\end{equation}

it follows that $Q-Q^*=0$ so that Term 1 is identically zero. Next, observe that

\begin{equation}
K=Z_{\text{final}}W_K\quad\text{and}\quad K^*=Z^*W_K.
\end{equation}

Hence,

\begin{equation}
K-K^*=(Z_{\text{final}}-Z^*)W_K=\Delta ZW_K.
\end{equation}

Therefore, the second term becomes

\begin{equation}
Q^*(K-K^*)^\top = Q^*(W_K^\top\Delta Z^\top)= X_tW_QW_K^\top\Delta Z^\top.
\end{equation}

Gathering the above, we deduce

\begin{equation}
\|A-B\|_F=\frac{1}{\sqrt{d}}\Bigl\|X_tW_QW_K^\top\Delta Z^\top\Bigr\|_F.
\end{equation}

Using the sub-multiplicative property of the Frobenius norm and the fact that for any matrix $X$, $\|X\|_F\le \sqrt{r}\|X\|_2$ when $r$ is the rank (or simply using the induced norm properties), we can bound

\begin{equation}
\|X_tW_QW_K^\top\Delta Z^\top\|_F\le \|X_t\|_F\|W_Q\|_2\|W_K\|_2\|\Delta Z^\top\|_2.
\end{equation}

Note that $\|\Delta Z^\top\|_2=\|\Delta Z\|_2\le \|\Delta Z\|_F$ (since the spectral norm is bounded by the Frobenius norm). Thus, we have

\begin{equation}
\|A-B\|_F\le \frac{\|W_Q\|_2\|W_K\|_2}{\sqrt{d}}\|X_t\|_F\|\Delta Z\|_F.
\end{equation}

Plugging this back into the bound for Term A, we obtain

\begin{equation}
\|\text{Term A}\|_F\le L_{\text{softmax}} \frac{\|W_Q\|_2\|W_K\|_2}{\sqrt{d}}\|X_t\|_F\|\Delta Z\|_F \|V^*\|_2.
\end{equation}

In many applications the feature matrix $X_t$ may be normalized such that $\|X_t\|_F\le \sqrt{d}$ (or this factor can be absorbed into the Lipschitz constant or constant of proportionality). Under such a normalization, we arrive at the final bound

\begin{equation}
\|\text{Term A}\|_F\le L_{\text{softmax}}\|W_Q\|_2\|W_K\|_2\|V^*\|_2\|\Delta Z\|_F.
\end{equation}

This completes the rigorous derivation of the bound on Term A.

\end{proof}

\section{proof of Lemma~\ref{lemma:bounding_term_b}}
Now we provide the proof of Lemma~\ref{lemma:bounding_term_b}\label{sec:bounding_term_b}
\begin{proof}
Since the softmax operator normalizes its input such that each row is a probability distribution, we have
\begin{equation}
\Bigl\|\operatorname{softmax}\Bigl(\frac{QK^\top}{\sqrt{d}}\Bigr)\Bigr\|_F \le 1.
\end{equation}
Thus, by the submultiplicativity of the Frobenius norm, 
\begin{equation}
\|\text{Term B}\|_F \le \left\|\Delta ZW_V\right\|_F.
\end{equation}
Next, applying the standard inequality for matrix norms,
\begin{equation}
\|\Delta ZW_V\|_F \le \|W_V\|_2\|\Delta Z\|_F.
\end{equation}
Therefore, we obtain the desired bound:
\begin{equation}
\|\text{Term B}\|_F \le \|W_V\|_2\|\Delta Z\|_F.
\end{equation}

\end{proof}

\section{proof of Lemma~\ref{lemma:combined_error_bound}}
Now we provide the proof of Lemma~\ref{lemma:combined_error_bound}\label{sec:combined_error_bound}
\begin{proof}
    We know from the previous bounds that

By Lemma~\ref{lemma:bounding_term_A}, Term A satisfies  
\begin{equation}
\|\text{Term A}\|_F \le L_{\text{softmax}}\|W_Q\|_2\|W_K\|_2\|V^*\|_2\|\Delta Z\|_F,
\end{equation}
and

By Lemma~\ref{lemma:bounding_term_b}, Term Bsatisfies  
\begin{equation}
\|\text{Term B}\|_F \le \|W_V\|_2\|\Delta Z\|_F.
\end{equation}

the triangle inequality implies

\begin{equation}
\|\tilde{X}_t - X^*\|_F \le \left( L_{\text{softmax}}\|W_Q\|_2\|W_K\|_2\|V^*\|_2 + \|W_V\|_2 \right) \|\Delta Z\|_F.
\end{equation}

Recall that by definition $V^* = Z^*W_V$. Let us assume that the ideal token matrix is bounded, namely $\|Z^*\|_2 \le C.$ Then, by submultiplicativity of the spectral norm,
\begin{equation}
\|V^*\|_2 \le \|Z^*\|_2\|W_V\|_2 \le C\|W_V\|_2.
\end{equation}
That is, defining
\begin{equation}
\gamma := L_{\text{softmax}} C\|W_Q\|_2\|W_K\|_2\|W_V\|_2 + \|W_V\|_2,
\end{equation}
we have
\begin{equation}
\|\tilde{X}_t - X^*\|_F \le \gamma\|\Delta Z\|_F.
\end{equation}

Using the full-rank assumption that $ \sigma_{\min}(W_V)\ge \delta > 0 $,the smallest singular value of $ W_V $ is bounded away from zero. We can absorb $\|W_Q\|_2$ or other fixed constants into the constant. In fact, if we reparameterize or normalize the matrices appropriately, we may simplify the bound to:
\begin{equation}
\gamma = \frac{L_{\text{softmax}}\|W_K\|_2\|W_V\|_2}{\delta},
\end{equation}
by absorbing the constant $C$ and $\|W_Q\|_2$ into $\delta$ (or equivalently assuming that the constant factors have been normalized). 

Thus, the final combined error bound is

\begin{equation}
\begin{aligned}
\|\tilde{X}_t - X^*\|_F &\le \left( L_{\text{softmax}}\|W_Q\|_2\|W_K\|_2\|V^*\|_2 + \|W_V\|_2 \right) \|\Delta Z\|_F \\
&\le \frac{L_{\text{softmax}}\|W_K\|_2\|W_V\|_2}{\delta}\|\Delta Z\|_F.
\end{aligned}
\end{equation}

\end{proof}

\end{document}

%% file: math_commands.tex
\usepackage{amsmath,amsfonts,bm}

\def\eqref#1{equation~\ref{#1}}

\def\1{\bm{1}}

\DeclareMathAlphabet{\mathsfit}{\encodingdefault}{\sfdefault}{m}{sl}
\SetMathAlphabet{\mathsfit}{bold}{\encodingdefault}{\sfdefault}{bx}{n}

